\pdfoutput=1
\documentclass[10pt,twocolumn,letterpaper]{article}

\usepackage{wacv}
\usepackage{times}
\usepackage{epsfig}
\usepackage{graphicx}
\usepackage{amsmath}
\usepackage{amssymb}
\usepackage{booktabs}
\usepackage{breqn}
\usepackage{mathtools}
\usepackage{color,soul} 

\def\wacvPaperID{379} 

\wacvalgorithmstrack   

\wacvfinalcopy 

\ifwacvfinal
\usepackage[breaklinks=true,bookmarks=false]{hyperref}
\else
\usepackage[pagebackref=true,breaklinks=true,colorlinks,bookmarks=false]{hyperref}
\fi

\begin{document}

\title{COPE: End-to-end trainable Constant Runtime Object Pose Estimation}


\author{Stefan Thalhammer\\
Automation and Control Institute, TU Wien\\
{\tt\small thalhammer@acin.tuwien.ac.at}
\and
Timothy Patten\\
Robotics Institute, UTS Sydney\\
{\tt\small timothy.patten@uts.edu.au}
\and
Markus Vincze\\
Automation and Control Institute, TU Wien\\
{\tt\small vincze@acin.tuwien.ac.at}
}

\maketitle
\thispagestyle{empty}

\begin{abstract}
State-of-the-art object pose estimation handles multiple instances in a test image by using multi-model formulations: detection as a first stage and then separately trained networks per object for 2D-3D geometric correspondence prediction as a second stage. 
Poses are subsequently estimated using the Perspective-$n$-Points algorithm at runtime.
Unfortunately, multi-model formulations are slow and do not scale well with the number of object instances involved. 
Recent approaches show that direct 6D object pose estimation is feasible when derived from the aforementioned geometric correspondences.
We present an approach that learns an intermediate geometric representation of multiple objects to directly regress 6D poses of all instances in a test image.
The inherent end-to-end trainability overcomes the requirement of separately processing individual object instances. 
By calculating the mutual Intersection-over-Unions, pose hypotheses are clustered into distinct instances, which achieves negligible runtime overhead with respect to the number of object instances.
Results on multiple challenging standard datasets show that the pose estimation performance is superior to single-model state-of-the-art approaches despite being more than $\sim$35 times faster. We additionally provide an analysis showing real-time applicability ($>24$ fps) for images where more than $90$ object instances are present.
Further results show the advantage of supervising geometric correspondence-based object pose estimation with the 6D pose.
\end{abstract}

\section{Introduction}

Object pose estimation is a challenging problem for monocular computer vision despite being essential for many tasks such as augmented reality, object manipulation, scene understanding, autonomous driving and industrial inspection~\cite{hou2020mobilepose,Nie_2020_CVPR,patten2020dgcm}.
Learning-based object pose estimation research focuses on maximizing the performance under challenging conditions like domain shift, object occlusion and object symmetries by tendentiously separating the detection from the pose correspondence estimation stage~\cite{haugaard2021surfemb,labbe2020cosypose,li2019cdpn,Park_2019_ICCV,sundermeyer2018implicit} then deriving 
the 6D pose with the Perspective-$n$-Points (P$n$P) algorithm~\cite{hartley2003multiple} using the estimated geometric correspondences.
This leads to shortcomings because a) adopting surrogate training targets decouples pose estimation from the training process and thus limits learning~\cite{di2021so} and b) running inference for multi-instance scenarios leads to a computational complexity of at least $\mathcal{O}\left(n\right)$  with respect to the number of objects ($n$) for the pose estimation stage.
Thus, this type of approach has severely diminishing applicability for realistic scenarios.

\begin{figure}[t]
   \centering
   \includegraphics[width=\columnwidth]{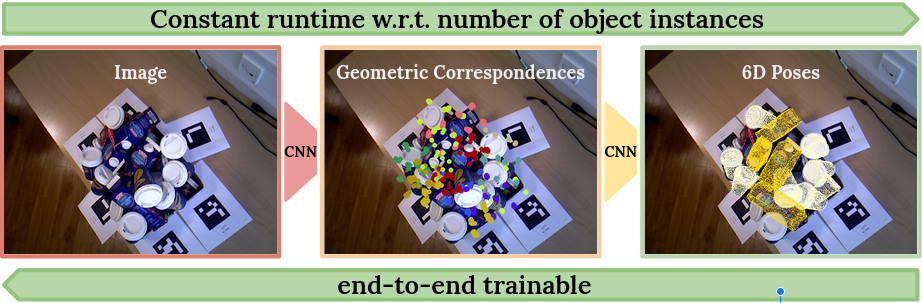}
   \caption{\textbf{COPE: Multi Object Instance 6D Pose Estimation with Constant Runtime.} Our end-to-end trainable pose estimator uses intermediate geometric correspondences to directly estimate 6D object poses from a single RGB image, thus processing multiple instances in parallel.}
   \label{fig:abov}
   \vspace{-2ex}
\end{figure}

Recent object pose estimation research trends recognize those shortcomings and partially alleviate them by directly regressing the 6D pose from the intermediate pose correspondences to achieve tremendous results~\cite{chen2020end,di2021so,hu2020single,wang2021gdr}.
In~\cite{wang2021gdr} and~\cite{di2021so}, detection is separated from the pose estimation stage, which makes them not end-to-end trainable because they require an object detector. The work in~\cite{hu2020single} is end-to-end trainable but separate networks need to be trained for each object and pooling geometric correspondences means multiple objects and instances cannot be handled simultaneously.
We improve over these findings by proposing a natural extension to efficiently handle multi-object multi-instance scenarios.

In this work we propose a solution to the aforementioned shortcomings by sharing the latent representation as well as the direct pose regressor over objects and their instances; see Figure~\ref{fig:abov} for a high-level overview.
We classify image locations in the feature maps, regress bounding box and view-dependent object geometry correspondences and regress the direct 6D pose.
While the first three of these tasks are intermediate representations, the direct 6D pose head is an up-stream task shared over those intermediate outputs of the network. 
Consequently, the loss related to the 6D pose is also backpropagated to the down-stream task of geometric correspondence estimation.
This design also allows further guidance of the learning process by enforcing consistency between these consecutive tasks, which additionally improves each of them.
We propose a concurrent solution to anchors~\cite{ren2015faster} for true location sampling during training that does not require manually choosing hyperparameters based on the expected test data distribution.
True locations are sampled and regression targets are standardized from a scalar shape prior derived from the respective object mesh and the backpropagated loss is normalized for each object class.
Thus, training is not biased towards larger objects and no prior assumptions need to be made in contrast to the case for anchors.

In summary, our contributions are:
\begin{itemize}
    \item A simple and efficient solution for multi-object multi-instance object pose estimation that improves over the state of the art.
    \item A training target sampling scheme that requires no assumptions about the test data distribution.  
\end{itemize}
 
Efficiently sharing internal representations over objects and instances enables end-to-end trainability that requires only one forward pass through the network to process all object instances in a single input image. 
We show that processing more than $90$ object instances in a single image with more than $24 fps$ on a modern consumer GPU, our method's performance is competitive to similar state-of-the-art approaches but up to $35$ times faster.

The remainder of the paper discusses related work in Section~\ref{sec:related_work}, followed by a description of our proposed approach in Section~\ref{sec:cope} and evaluations in Section~\ref{sec:Experiments}. Lastly, Section~\ref{sec:conclusion} concludes the paper.

\section{Related Work} \label{sec:related_work}

In this section we present the state of the art for monocular 6D object pose estimation with a focus on work that directly regress the object pose.
This is followed by reviewing training target sampling for object detection.

\textbf{Object Pose Estimation}
Since direct pose regression from feature space leads to inferior performance, the dominant monocular object pose estimation approaches leverage geometric correspondences as regression targets~\cite{hodan2020epos,li2019cdpn,oberweger2018making,Park_2019_ICCV,park2020latentfusion,hu2019segpose,zakharov2019dpod}.
Poses are derived for each estimated set of object correspondences using variants of P$n$P~\cite{hartley2003multiple,lepetit2009epnp,terzakis2020consistently}.
Recent trends replace the classical solver with trainable versions~\cite{chen2020end,di2021so,hu2020single,song2020hybridpose,wang2021gdr} to infer the 6D pose directly from the intermediate geometric correspondences.
This enables end-to-end trainable object pose estimation as it provides the additional supervision for the down-stream network parts with the 6D pose.
Their findings indicate that direct 6D pose estimation also results in state-of-the-art performance by sharing the pose regressor over objects~\cite{di2021so,wang2021gdr}.
However, efficient and simultaneous single-stage multi-object instances handling is a problem that still remains~\cite{hodan2020epos}.

Considering the top performing approaches in the BOP challenge~\cite{hodan2020bop}, a benchmark that aims to provide a standardized protocol for an unbiased comparison of object pose estimation, a frequently used technique to handle multiple object instances in an image is to separate object detection from pose estimation~\cite{li2019cdpn,labbe2020cosypose,hodan2020epos,liu2020leaping,Park_2019_ICCV,haugaard2021surfemb}.
In the first stage, 2D location hypotheses are provided using common object detectors such as  Faster-RCNN~\cite{ren2015faster}, RetinaNet~\cite{lin2017focal} or FCOS~\cite{tian_fcos}.
In the second stage, object crops are passed to the pose estimator but this leads to considerable temporal and computational cost.
An exception is EPOS~\cite{hodan2020epos} where multi-instance handling is facilitated by using Graph-Cut RANSAC~\cite{GCransac} to cluster the predicted geometric correspondences to individual instances. 
Despite providing a sophisticated approach for addressing object symmetries and multiple object instances with one forward pass through a network, their multi-instance fitting of poses using~\cite{GCransac} is computationally very demanding.
In our work, we alleviate this issue by adopting ideas from object detectors and incorporate direct pose regression into the detection stage.

\textbf{Detecting Objects in Images}
Single-stage object detectors offer efficient solutions for multi-object and multi-instance object localization in 2D~\cite{bochkovskiy2020yolov4,duan2019centernet,law2018cornernet,liu2016ssd,lin2017focal,tian_fcos}.
Object detection uses anchors~\cite{lin2017focal,liu2016ssd,ren2015faster} to sample bounding box priors with different sizes and aspect ratios in the multi-scale feature map of feature pyramids~\cite{lin2017feature}.
For training, foreground image locations are chosen based on the Intersection-over-Union (IoU) between ground truth bounding boxes and the anchor boxes.
As such, training locations are correlated with the projected object shape in the image space.
This leads to effective encoding and handling of objects with different scales.

Using anchors has two downsides.
Firstly, it requires the manual specification of $16$ hyperparameters that reflect the expected training and test dataset statistics.
Secondly, the size of the output space depends on the number of anchors sampled for each image location. 
Recent approaches propose alternative formulations to circumvent these shortcomings while retaining the advantages of anchors~\cite{duan2019centernet,law2018cornernet,tian_fcos,zhang2019freeanchor}.
The authors of~\cite{tian_fcos} choose the respective feature map resolution for training explicitly by using the bounding box size to overcome the necessity for sampling anchor boxes.
True image locations of the feature map, for loss backpropagation, are assigned based on the respective pixel's centerness with respect to the ground truth bounding box.
Alternatively,~\cite{law2018cornernet} models objects as paired-keypoints: the top-left and the bottom-right corner of the bounding box.
Similarly,~\cite{duan2019centernet} addresses the inefficiency of anchor-based object detection by modeling objects as their center points and estimating the bounding box relative to it.
The authors of~\cite{zhang2019freeanchor} overcome the requirement of hyperparameters for assigning objects to anchors by designing a flexible maximum likelihood estimation assignment for network training.

We propose to sample training locations based on the visible object mask and the 3D object dimensions. 
We also use the 3D object dimension to effectively replace the anchor-based target annotation standardization.
As such, we encode objects of different sizes and eccentricities more effectively, while also reducing the size of the output space and the number of required hyperparameters.

\section{Constant Runtime Object Pose Estimation} \label{sec:cope}

This section provides an overview of the proposed direct 6D pose estimation framework, which we name \textbf{C}onstant Runtime \textbf{O}bject \textbf{P}ose \textbf{E}stimation, abbreviated as COPE.
We start with a high-level overview of the method.
Afterwards, we detail the approaches for deriving anchor-free training targets, true image location sampling and geometry correspondence standardization during training.
This is followed by an explanation of how the direct 6D pose is parameterized and how symmetries are handled.
We conclude with a description of multi-instance clustering and hypotheses filtering during testing.

\subsection{Constant Runtime via Direct-pose regression}

\begin{figure*}[t]
   \centering
   \includegraphics[width=\textwidth,height=7.2cm]{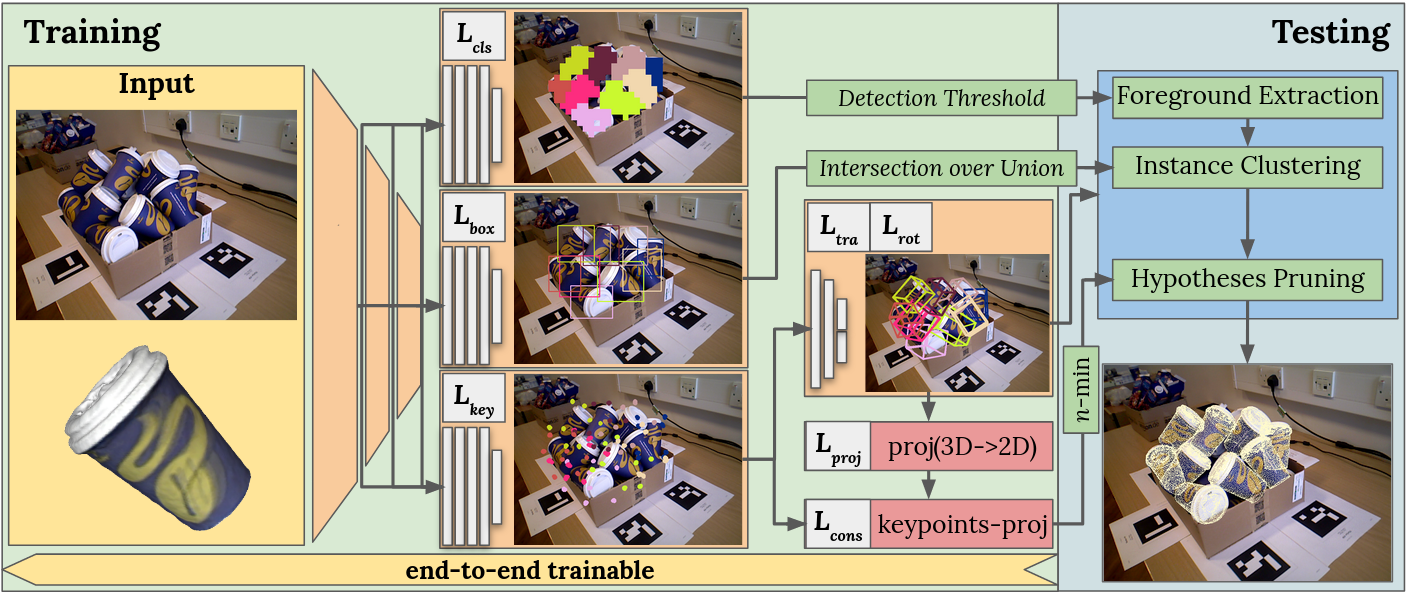}
   \caption{\textbf{Constant Runtime Object Pose Estimation.} Given an input image and a 3D model, image locations are classified while bounding boxes and geometric correspondences are regressed. 
   A direct pose regression module slides over the image locations and regresses the 6D pose from the geometric correspondences. Training is supervised with losses for each module ($L_{cls}$, $L_{box}$, $L_{key}$, $L_{tra}$ and $L_{rot}$) as well as auxiliary losses ($L_{proj}$ and $L_{cons}$) to enforce consistency between estimated correspondences and direct poses. During testing, instances are efficiently clustered using their 2D IoU then the $n$ hypotheses with the highest consistency generate the 6D output.}
   \label{fig:COPE}
\end{figure*}

Our aim is to classify and estimate the poses of all object instances in a single RGB input image. 
The 6D pose is defined as $\hat{P} \in SE(3)$, which represents the object's rotation $R \in \mathbb{R}^{3}$ and translation  $t \in \mathbb{R}^{3}$ with respect to the camera's coordinate frame.
Object meshes are considered to be known in advance but no additional information regarding the test scene is required.
We define the corner points of the smallest cuboid enclosing the respective object mesh in its coordinate frame as the geometric correspondences~($G_{3D}$).
COPE, outlined in Figure~\ref{fig:COPE}, outputs the set of object instances visible in the image, parameterized with object type and 6D pose.

COPE builds upon the success of recent efficient object detection approaches~\cite{lin2017focal,tian_fcos,zhang2020bridge}. 
The RGB input image is first processed with a CNN backbone and then multi-scale features are computed using a feature pyramid to estimate the intermediate object representation. 
Three modules shared over feature maps of sizes [$s/8$, $s/16$, $s/32$], with $s$ being the input image resolution, generate the intermediate outputs $\hat{O}$, $\hat{B}$ and $\hat{G}$.
The first module predicts the set of object class probabilities $\hat{O} \coloneqq \{\ \hat{o}_{0}, ..., \hat{o}_{k}\}$, where $k$ is the number of image locations in the multi-scale feature map and $\hat{o}_{k} \in \mathbb{R}^{a}$ is the Bernoulli-distributed object class prediction.
We denote the number of object classes in the dataset with $a$.
The second module predicts the amodal bounding boxes $\hat{B} \coloneqq \{\ \hat{b}_{0}, ..., \hat{b}_{k}\}$, where $\hat{b}_{k} \in \mathbb{R}^{4}$.
The third module predicts the projection of $G_{3D}$ in the image space $\hat{G} \coloneqq \{\ \hat{g}_{0}, ..., \hat{g}_{k}\}$, where $\hat{g}_{k} \in \mathbb{R}^{16}$ represents the 2D coordinates of the $8$ corner points of the smallest cuboid enclosing the respective object mesh. 
A shared direct pose module slides over the set $\hat{G}$, estimating direct pose hypotheses $\hat{P} \coloneqq \{\ \hat{p}_{0}, ..., \hat{p}_{k}\}$, where $\hat{p}_{k} \in \mathbb{R}^{9}$.
The pose output is parameterized by $3$ values for translation in $\mathbb{R}^{3}$ and the first two basis vectors of the rotation matrix in $\mathbb{R}^{3}$, thus $6$ values in total~\cite{zhou2019continuity}.
A set $\hat{C} \coloneqq \{\ \hat{c}_{0}, ..., \hat{c}_{k}\}$ is computed to quantify the consistency $\hat{c}_{k} \in \mathbb{R}^{1}$ between $\hat{g}_{k}$ and $proj_{3D\to2D}(G_{3D}\cdot\hat{p}_{k})$ for each image location separately.

During inference, the network predicts $H = \{\hat{O}$, $\hat{B}$, $\hat{G}$, $\hat{P}$, $\hat{C} \}$ with constant runtime for a given query image.
Corresponding elements of $H$ with an image location $k$ of maximum class probability $\hat{o}_{k}$ below the detection threshold are discarded.
The resulting subsets are clustered into object instances using the IoU between elements of $\hat{B}$.
We define a hyperparameter $n$ for the highest number of consistencies in $\hat{C}$.
This parameter is set to $10$ for the presented experiments; see the ablation in Table~\ref{tab:DPR}.
Finally, the detected object classes and the mean of the $n$ poses with the highest consistencies per instance are returned with negligible increase in runtime with respect to the number of object instances. Through this procedure, our method can estimate the poses of a large number of object instances in a single test image in real-time ($> 24 fps$) on an Nvidia Geforce 3090 GPU. 


\subsection{Training Target Sampling}

Effective assignment of true image locations for updating network weights during training is an ongoing research problem~\cite{duan2019centernet,law2018cornernet,lin2017focal,tian_fcos,zhang2020bridge,zhang2019freeanchor}.
These true image locations are often sampled in the output feature maps of feature pyramids~\cite{lin2017feature}, which are a great tool to efficiently encode scale information in the feature space.
Anchors~\cite{ren2015faster} are the standard representation for providing bounding box priors to sample true image location based on the IoU with the ground truth during training~\cite{ren2015faster,liu2016ssd,lin2017focal}.
For each image location in the multi-scale feature map, $9$ differently shaped and sized bounding box priors are sampled. This requires $16$ hyperparameters: $5$ each for \textit{base sizes} and \textit{strides} and $3$ each for \textit{ratios} and \textit{scales}~\cite{lin2017focal}.
This results in two convenient traits since anchor locations used for updating the network's weights are chosen based on a threshold parameter for the IoU with the ground truth:
\begin{itemize}
    \item Sampling anchors leads to a uniform scale space for the expected bounding boxes.
    As such, a similar amount of training locations are sampled per object, independent of the object's size in the image space.
    \item Regression targets are standardized using the respective anchor's center, width and height. This means that the regression target space has similar statistics for differently sized objects. 
\end{itemize}
Despite these convenient traits, training target sampling can still be improved since anchors require a) choosing $16$ hyperparameters depending on the expected object scales in image space and b) generating $9$ anchors per feature map location, which results in a large output space that slows down convergence.
We overcome these shortcomings by using a regression target standardization scheme that reflects the object's geometry and scale.

\subsubsection{True Location Sampling}
Object masks are used for true training location sampling as in~\cite{hu2019segpose}. However, instead of predicting object masks and correspondences from a single feature map resolution, our work adopts the divide-and-conquer strategy of feature pyramids to make predictions from multiple feature map resolutions.
To overcome the necessity of requiring hyperparameters~\cite{ren2015faster,tian_fcos} for choosing the best suited feature map resolution for locating an object, we propose a geometry-based approach to assign true training locations.
We supplement true location sampling with a scalar shape prior:
\begin{align} \label{eq:diameter}
    \delta_{o}={max} ||(m_{i} - m_{j})||_{2} \hspace{1ex} \forall \hspace{1ex} m_i, m_j \in M, i \neq j
\end{align}
where $M$ is the set of object model vertices. 
Since the spatial downsampling of the input image through the backbone follows an exponential function, it is intuitive to explicitly choose pyramid levels using a logarithmic function.
As such, we choose the respective feature pyramid level with:
\begin{align} \label{eq:location_sampling}
    level = f + log_{d} (\delta_{o} / t_{z}),
\end{align}
where $f$ depends on the number of pyramid levels used, $t_{z}$ corresponds to the object's distance from the camera and $d$ is the only remaining hyperparameter.
Since we use three pyramid levels, as in~\cite{thalhammer_pyrapose}, this requires choosing only $6$ hyperparameters for FCOS and $12$ when using anchors. 
An additional advantage is that $\delta_{o}$ better reflects the object shape in all three spatial dimensions and thus also the visible object surface in the image space compared to using the bounding boxes for the assignment of true training locations.
As a consequence, elongated objects are tendentiously sampled in higher resolved feature pyramid levels than boxy shaped objects.
Despite needing fewer hyperparameters, we retain a similar amount of true locations used for training.
Classifying true image locations ($L_{cls}$) is supervised using the focal loss~\cite{lin2017focal}.

\subsubsection{Geometric Correspondence Standardization} 

Instead of standardizing the projected object correspondences $G$ using anchor priors or with a scalar value agnostic to object shape~\cite{tian_fcos}, we directly incorporate $\delta_{o}$ to scale regression targets of different objects to a similar magnitude:
\begin{align}\label{eq:key_stand}
  y_{G} = (c - G) / \delta_{o},
\end{align}
where $c$ is the center of the respective feature map location, $G$ are the image locations of the geometric correspondences and $y_{G}$ are the standardized regression targets.
As such, regression targets are encoded similarly as with anchors (with similar $\sigma$ for $G$ for all objects independent of their scale or shape eccentricity).
Thus, the computed error is independent of the object's scale in the image space and the training process is not biased for larger objects.
Our approach needs no hyperparameters for standardization and convergence is improved since $9$ times fewer network output parameters per feature map location are required compared to anchors.


\subsubsection{Imbalance Problem of Target Locations}

Choosing training target locations based on the object mask leads to a training process that is biased towards objects with a larger projected image surface.
For classification this is commonly circumvented using the focal loss~\cite{ren2015faster}.
Using anchors as location priors alleviates the issue since anchors are sampled uniformly over the expected object scale space.
We define a concurrent solution by normalizing over the number of true training locations $l$ and accumulating the gradient afterwards. 
The regression loss is:
\begin{align}\label{eq:loss_reg}
  L_{reg}(\hat{y},y) = \dfrac{1}{a} \cdot \sum_{i=0}^{a} \dfrac{1}{l_{i}} \cdot \sum_{j=0}^{l_{i}} huber(\hat{y}_{j}, y_{j}),
\end{align}
where $huber$ is the augmented $l_{1}$ loss used in RetinaNet~\cite{hastie2009elements,lin2017focal} and $y$ and $\hat{y}$ are the ground truth and estimate, respectively.
This procedure requires no additional trainable parameters and only leads to minor computational overhead during training time and to none during test time despite improving multi-object handling. 

\subsubsection{Direct Pose Regression}
The direct pose is regressed using the output $\hat{y}$ of the module estimating intermediate geometric correspondences as done in~\cite{di2021so,hu2020single,song2020hybridpose,wang2021gdr}.
The 6D pose is parameterized as $P\in SE(3)$, with $t \in \mathbb{R}^{3}$ being the 3D translation vector and $R \in \mathbb{R}^{6}$ the first two base vectors of the $SO(3)$ rotation matrix as done in~\cite{di2021so,kundu20183d,wang2021gdr}.

Prior methods perform pose estimation on zoomed crops of the detected objects of interest.
Using the object rotation in the camera coordinate system, i.e. the allocentric rotation~\cite{kundu20183d}, leads to mapping different object views to the same rotation.
To alleviate that problem, these approaches use the rotation of the camera in the object coordinate frame, i.e. the egocentric rotation.

In contrast, we learn to predict geometric correspondences directly in the image space.
These correspondences are destandardized with the inversion of Equation~\eqref{eq:key_stand} and fed to the direct pose estimation module.
As such, our approach correlates object rotation with its image location. 
This means that we are able to directly regress the allocentric rotation since we require no zooming or cropping.
Additionally, we can directly regress the 3D translation without requiring a scale-invariant translation representation as used in~\cite{di2021so, li2019cdpn, wang2021gdr}.
The network training is supervised using the image locations sampled with Equation~\eqref{eq:location_sampling}.



\subsection{Symmetry-aware Loss}

Objects exhibiting discrete or continuous symmetries, i.e. similar views that correspond to different ground truth poses \textit{P}, are detrimental to the convergence of the network training~\cite{manhardt2019explaining,Park_2019_ICCV,shi2021stablepose}.
We adopt the transformer loss of~\cite{Park_2019_ICCV} since symmetries are efficiently handled during loss computation and require no additional trainable weights.
We define our keypoint estimation loss for supervising the training of the geometric correspondence learning with: 
\begin{align}\label{eq:key_pos}
    L_{key} = \underset{s \in S_{i}}{min}\ L_{reg}(\hat{y}, hy),
\end{align}
where $S_{i}$ is a set of symmetry transformations that depend on the visual ambiguities of the object.
We observed that separately choosing hypotheses with \textit{L\textsubscript{key}}, and substituting the direct pose losses for \textit{L\textsubscript{rot}} and \textit{L\textsubscript{tra}} with Equation~\eqref{eq:key_pos} introduces ambiguities since the 6D pose is directly derived from the estimated intermediate geometric correspondences.
To alleviate this issue we define an indicator function $\mathbb{I}$, indicating the symmetry that minimizes $L_{key}$.
As such, we supervise the direct pose regression with:
\begin{align}\label{eq:sym_pos}
    L_{rot / tra} = L_{reg}(\hat{y}, \mathbb{I}(S)y).
\end{align}
Since only one set of $\hat{G}$ is predicted per image location and Equations~\eqref{eq:key_pos} and~\eqref{eq:sym_pos} sufficiently account for object symmetries, $L_{proj}$ and $L_{cons}$ can be directly computed with $L_{key}$.
The projection and the consistency loss are thus defined as:
\begin{align}
    L_{proj} = L_{reg}(G_{3D}\hat{P}, G), \\
    L_{cons} = L_{reg}(G_{3D}\hat{P}, \hat{G}).
\end{align}
The overall loss is:
\begin{multline}\label{eq:dir_pos}
    L = \alpha \cdot L_{cls} + \beta \cdot L_{box} + \gamma \cdot L_{key} + \delta \cdot L_{rot} \\ + \epsilon \cdot L_{tra} + \zeta \cdot L_{proj} + \eta \cdot L_{cons},
\end{multline}
where $\alpha$, $\beta$, $\gamma$, $\delta$, $\epsilon$, $\zeta$ and $\eta$ are loss weights. The bounding box estimation, $L_{box}$, is supervised using Equation~\eqref{eq:loss_reg}.

\subsection{Multi-instance Handling}

Commonly, multiple instances of the same object in a single image are handled before correspondence estimation by non-maximum suppression of the detection stage~\cite{li2019cdpn,wang2020self6d,labbe2020cosypose,Park_2019_ICCV} or by clustering correspondences afterwards~\cite{hodan2020epos}.
The first family of methods individually process each instance's image crop to estimate the 6D pose. 
The second family of methods is more advantageous because the network is shared over all objects of interest.
Unfortunately, since~\cite{hodan2020epos} predicts dense geometric correspondences the method has a high runtime. This is due to the clustering of correspondences to object instances using~\cite{GCransac}, which is computationally demanding.

In our work, $H$ is first filtered by discarding the non-maximally scoring object classes for each image location $k$. 
Subsequently, image locations with a detection score below the detection threshold are pruned.
The remaining hypotheses correspond to detected objects.
The 2D bounding boxes, $\hat{B}$, are used to cluster object instances based on the respective IoU between the outputs of different image locations.
Ultimately, using the computed consistency $\hat{C}$, the pose is averaged over the $n$ hypotheses of $\hat{P}$ with the highest consistency for each object instance.



\section{Experiments} \label{sec:Experiments}

This section provides quantitative and qualitative evaluations of COPE on several datasets.
After introducing the experimental setup, we proceed with comparisons to the state of the art on two challenging datasets using the BOP protocol~\cite{hodan2020bop}.
In addition, ablation studies are presented to quantify the influence of direct pose supervision on an additional dataset.
To further validate and thoroughly test the capabilities of our method, we present results on a synthetic dataset with up to $100$ object instances per image.



\subsection{Datasets} \label{sec:Datasets}

Evaluation is provided on three standard datasets: LM~\cite{hinterstoisser2012model}, LM-O~\cite{brachmann2014learning} and IC-BIN~\cite{DBLP:journals/corr/DoumanoglouKMK15}.
For evaluation, we use the subsets provided with the BOP challenge. LM provides $200$ test images for each of the $13$ objects that come with watertight object models.
LM constitutes a common benchmark for object pose estimation in cluttered environments.
LM-O consists of $200$ test images of LM's second test sequence with all eight objects annotated in each. 
LM-O provides test images with challenging object occlusions.
IC-BIN presents $150$ test images of up to $21$ instances of two objects with heavy occlusion.

For training we use the $50$k photorealistic renderings for each dataset available through the BOP challenge~\cite{denninger}. These are generated using physically based rendering (\textit{pbr})~\cite{hodavn2019photorealistic}.
Results on LM and LM-O are provided with the same models trained on all $13$ objects of LM.
No annotated real images are used for training.

\subsection{Evaluation Metrics}

Comparison to the state of the art is provided using the performance score of the BOP challenge~\cite{hodan2020bop}.
Results for pose estimation are reported using the Average Recall: $AR = (AR_{VSD} + AR_{MSSD} + AR_{MSPD})/3$.
Ablations are evaluated using the ADD recall, or ADD-S recall for objects exhibiting symmetries~\cite{hinterstoisser2012model}.
We report the fraction of poses below the commonly used error threshold of $10\%$ of the object diameter.
Results for object detection are reported using the the mean Average Precision (mAP) of the Microsoft COCO object detection challenge~\cite{lin2014microsoft}.
The results are those for the IoU values from $0.5$ to $0.95$ in $0.05$ steps. 
Please refer to the supplementary material for details.

\subsection{Implementation Details}


The weights of the backbone are pre-trained on ImageNet~\cite{russakovsky2015imagenet} and fine-tuned for $120$ epochs using the Adam~\cite{kingma2014adam} optimizer with a learning rate of $10^{-5}$ and a batch size of $8$. 
Previous work suggests overcoming the domain gap between training on synthetic and testing on real images by not updating certain network weights during optimization~\cite{hinterstoisser_pretraining,zakharov2019dpod}.
Similarly, we do not update parameters of batch normalizations and the convolutions of the first two stages of the backbone during fine-tuning.
We also apply image augmentations as described in~\cite{thalhammer_pyrapose}.
The parameter $d$ in Equation~\eqref{eq:location_sampling} is set to $3$ for all experiments.

\subsection{Comparison to the State of the Art}

\textbf{Object Pose Estimation} This section compares the performance of COPE to the state of the art on IC-BIN and LM-O.
Results using the BOP setting reporting the \textit{AR} are provided in Table~\ref{tab:SOA}.
The bottom section compares single-model methods, i.e approaches that produce estimates for all object classes and their instances in a single forward pass. 
Both DPOD~\cite{zakharov2019dpod} and EPOS~\cite{hodan2020epos} require P\textit{n}P for deriving the 6D pose from the predicted geometric correspondences, while COPE directly outputs the 6D pose.
COPE improves over both methods in \textit{AR} on average.
Compared to the previous single-model state of the art, EPOS, COPE achieves similar \textit{AR} on LM-O $0.543$ as compared to $0.547$ but improves to $0.440$ in comparison to $0.363$ on IC-BIN.
More remarkable, however, is that the runtime of COPE is $37$ times faster using the inference speed calculated by the BOP toolkit\footnote{https://github.com/thodan/bop\_toolkit}.

The top section of Table~\ref{tab:SOA} presents the results of multi-model methods.
These multi-model methods use an object detector to sample sparse location priors then separately trained networks per object class for correspondence prediction and pose estimation.
For the methods~\cite{di2021so,hu2022perspective,lipson2022coupled,su2022zebrapose} no results on IC-BIN are available.
Compared to the best individually performing methods on both datasets, CosyPose on IC-BIN and ZebraPose on LM-O, COPE results in $\sim24\%$ relative performance decrease.
This is in the expected range due to the known performance decrease for single-staged approaches~\cite{lin2017focal}.

\begin{figure}

    \centering
    \includegraphics[width=0.95\linewidth]{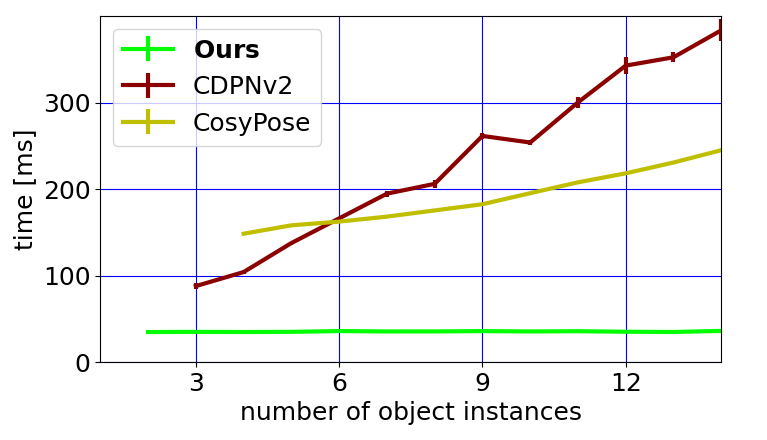}
    \caption{\textbf{Runtime Comparison to the State of the Art on IC-BIN.} Provided are the times it takes to estimate poses for all object instances in a single image.}
    \label{fig:runtime}


\end{figure}

\textbf{Runtime} Figure~\ref{fig:runtime} presents the average runtime and standard deviation on IC-BIN for five test runs of COPE, CDPNv2 and CosyPose on an Intel CPU with 3.6GHz and an Nvidia Geforce 3090 GPU. 
The times reported for CDPNv2 exclude the time required for detecting objects.
Despite omitting the runtime of CDPNv2's first stage, our method is more than $12$ times faster and $7$ times faster than CosyPose when processing $15$ object instances.
Most notably, in contrast to multi-model approaches, COPE is capable of directly providing 6D poses for multi-object multi-instance cases at almost constant runtime, which makes it highly suitable for real-time scenarios.

\begin{table}[t!]
    \centering
    \caption{\textbf{Comparison to the State of the Art for Pose Estimation.} Presented are the \textit{Average Recall} on IC-BIN and LM-O, the average over both and the inference speed using the BOP toolkit.}
    \label{tab:SOA}

\begin{tabular}{c|c|c||c|c}
\hline
Method & IC-BIN & LM-O & Avg. & Time \\ \hline \hline
\multicolumn{5}{c}{\textit{\textbf{Multi-model}}} \\ \hline
AAE~\cite{sundermeyer2018implicit} & 0.217 & 0.146 & 0.182 & \textbf{0.199} \\
Pix2Pose~\cite{Park_2019_ICCV} & 0.226 &  0.363 & 0.295 & 1.230 \\
2Dto3D~\cite{liu2020leaping} & 0.342 &  0.525 & 0.434 & 0.546  \\ 
CDPNv2~\cite{li2019cdpn} & 0.473 & 0.624 & 0.549 & 1.010 \\
SurfEmb~\cite{haugaard2021surfemb} & 0.550 & 0.623 & 0.587 & 6.296\\ 
CosyPose~\cite{labbe2020cosypose} & \textbf{0.574} & 0.618 & 0.\textbf{596} & 0.227\\
SO-Pose~\cite{di2021so} & - & 0.613 & - & -  \\
CIR~\cite{lipson2022coupled} & - & 0.655 & - & -  \\
PFA~\cite{hu2022perspective} & - & 0.683 & - & - \\
ZebraPose~\cite{su2022zebrapose} & - & \textbf{0.718} & - & 0.250 \\ \hline \hline
\multicolumn{5}{c}{\textit{\textbf{Single-model}}} \\ \hline
DPOD~\cite{zakharov2019dpod} & 0.169 & 0.130 & 0.150 & 0.211 \\
EPOS~\cite{hodan2020epos} & 0.363 &  \textbf{0.547} & 0.455 & 2.804 \\
\textbf{Ours} & \textbf{0.440} &  0.543 & \textbf{0.492} & \textbf{0.075}\\ \hline
\end{tabular}
\end{table}

\textbf{Object Detection} Table~\ref{tab:SOA_det} compares the object detection accuracy (\textit{mAP}~\cite{lin2014microsoft}) of COPE to the state of the art on IC-BIN and LM-O using the same training data.
On average, COPE outperforms both MaskRCNN~\cite{he2017mask} and FCOS~\cite{tian_fcos}, achieving the highest average \textit{mAP} over both datasets.
On LM-O, COPE is superior to MaskRCNN, achieving $0.532$ as compared to $0.375$ but slightly inferior to FCOS that reaches $0.622$.
This is partly due to the lean network design of COPE.
Please refer to the supplementary material for more details.
COPE's detected bounding boxes are more precise than both standard detectors used by many multi-model methods on IC-BIN, achieving $0.431$ as compared to $0.323$ and $0.316$. 
As such our method provides excellent location priors for pose refinement.

\begin{table}[t!]
    \centering
    \caption{\textbf{Comparison to the State of the Art for Object Detection.} Presented are the \textit{Average Recall} on IC-BIN and LM-O and the average using the BOP toolkit.}
    \label{tab:SOA_det}
\begin{tabular}{c|c|c||c}
\hline
Method & IC-BIN & LM-O & Avg.\\ \hline
MaskRCNN~\cite{he2017mask,labbe2020cosypose,haugaard2021surfemb} & 0.316 & 0.375 & 0.346 \\
FCOS~\cite{tian_fcos, li2019cdpn, su2022zebrapose} & 0.323 & \textbf{0.622} & 0.473 \\
\textbf{Ours} & \textbf{0.431} & 0.532 & \textbf{0.482} \\ \hline
\end{tabular}
\end{table}

\subsection{Ablation Studies}


\textbf{Runtime Evaluation}
In order to exhaustively test the runtime and scalability of COPE, we create a synthetic test dataset using the IC-BIN objects and OpenGL\footnote{https://github.com/thodan/bop\_renderer} rendering, named \textit{IC-BIN syn}.
The number of object instances to render per image is sampled from a uniform distribution with a lower bound of $10$ and upper bound of $100$.
We render the sampled number of object instances randomly from the IC-BIN objects onto the test images of IC-BIN.
Results are again provided for processing one test image on an Intel CPU with 3.6GHz and an Nvidia Geforce 3090 GPU.

\begin{figure}[!t]
   \centering
   \includegraphics[width=\linewidth]{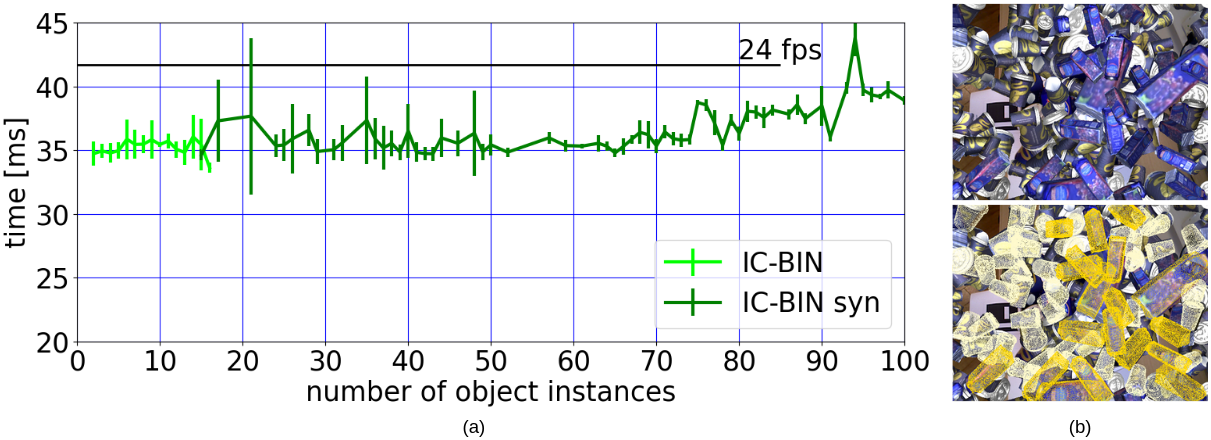}
   \caption{\textbf{Runtime Evaluation of COPE on IC-BIN and IC-BIN syn.} (a) COPE provides negligible runtime increase up to $70$ object instances per image. The black line indicates the threshold for real-time processing. (b) An example of a test image of \textit{IC-BIN syn} and a visualization of the estimated poses.}
   \label{fig:ludicrous_runtime}
\end{figure}

Figure~\ref{fig:ludicrous_runtime} (a) presents the runtime of our method for detecting and estimating the poses of up to $100$ object instances in a single image.
We report the average runtime and standard deviation for five test runs.
The runtime increases negligibly up to $70$ detected object instances.
For more than $90$ instances per image, our method exceeds real-time processing.
As such, it provides quantitative proof of the tremendous scalability of the presented approach and the constancy of the runtime with respect to the number of object instances in a single test image.
Figure 4 (b) provides a rendered synthetic test image (top) and projected object models based on the estimated poses (bottom).

\textbf{Direct-pose Regression}
Table~\ref{tab:DPR} displays the influence of direct pose regression on the end-to-end architecture on the LM~\cite{hinterstoisser2012model} and LM-O~\cite{brachmann2014learning} datasets using 
the \textit{ADD/(-S)} recall.
The column \textit{Voting} indicates the pose voting procedure using RANSAC-EP\textit{n}P, an average of all direct pose hypotheses, or an average of the direct pose estimates with the best \textit{n} hypotheses in terms of $\hat{C}$.

\begin{table}[!t]
\centering

\caption{\textbf{Ablation Study for Pose Supervision.} Provided is the average \textit{ADD/(-S)} recall. The objects \textit{Eggbox} and \textit{Glue} are considered as symmetric objects.}
\label{tab:DPR}

\begin{tabular}{c c|c|c|c} 
\hline
\multicolumn{2}{c|}{Supervision} & Voting & LM & LM-O \\ \hline
\textbf{IM} & 2D & PnP & 0.654 & 0.280 \\ \hline
\textbf{DR} & 2D & PnP & 0.712 & 0.330 \\ 
& 6D & all & 0.715 & 0.342 \\ \hline
\textbf{DR-P}  & 2D & PnP & 0.672 & 0.341 \\ 
 & 6D & all & 0.672 & 0.345 \\ \hline
\textbf{DR-PC} & 2D & PnP & 0.712 & 0.348 \\
 & 6D & n=1  & 0.722 & 0.338 \\ 
 & 6D & n=5 & 0.724 & 0.346 \\
 & 6D & n=10 & 0.732 & \textbf{0.350} \\
 & 6D & all & \textbf{0.738} & 0.349 \\ \hline
\end{tabular}

\end{table}

\begin{figure}[!t]
    \centering
    \includegraphics[width=\linewidth]{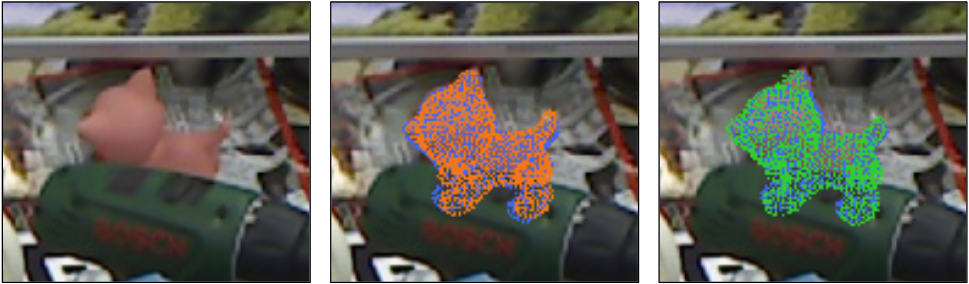}
    \caption{\textbf{Pose Supervision Comparison on LM-O's \textit{Cat}.} From left to right: raw image, pose obtained from geometric correspondences and RANSAC-EP\textit{n}P, and direct pose regression. Blue, red and green meshes indicate ground truth, false positive and true positive pose (as measured by \textit{ADD}).}
   \label{fig:DR_viz}
\end{figure}

The results show that supervising the training process with direct pose regression (\textbf{DR}) improves the quality of the intermediate representation (\textbf{IM}) tremendously.
The improvement is from $0.654$ to $0.715$ on LM and from $0.280$ to $0.342$ on LM-O.
Using \textbf{DR} direct pose estimates is superior to using those estimated by RANSAC-EP\textit{n}P.
Providing additional guidance with \textit{L\textsubscript{proj}} (\textbf{DR-P}) improves for the occluded scenario of LM-O but is detrimental for LM.
Ultimately, enforcing consistency between the internal representation and the correspondences projected to 2D using the regressed 6D pose with \textit{L\textsubscript{cons}} (\textbf{DR-PC}) leads to good results for direct regression and when using the intermediate representation on both datasets.
Figure~\ref{fig:DR_viz} shows an example of LM-O's cat under occlusion. The re-projected model using the ground truth is colored blue and the wrong estimate based on \textit{ADD}, using the intermediate representation and RANSAC-EP\textit{n}P, is colored red (middle image). Direct pose regression recovers from the incorrect intermediate representation, which is displayed in green (right image).


\subsection{Qualitative Evaluation}

Figure~\ref{fig:qual} visualizes results on LM-O, LM and IC-BIN. 
Displayed are projected object meshes based on the estimated pose in the top row and estimated bounding boxes in comparison to the respective ground truth in the bottom row. 
Green and red bounding boxes portray estimates and ground truth, respectively.
The left image pair indicates a common error for LM-O: a false negative detection of the \textit{Eggbox}. 
The right image pair shows that some of IC-BIN's instances of \textit{Juice} are difficult to detect while detecting \textit{Coffeecup} works well even under heavy occlusion if more than the lid is visible.

\setcounter{figure}{5}
\begin{figure}[!t]
   \centering
   \includegraphics[width=\linewidth]{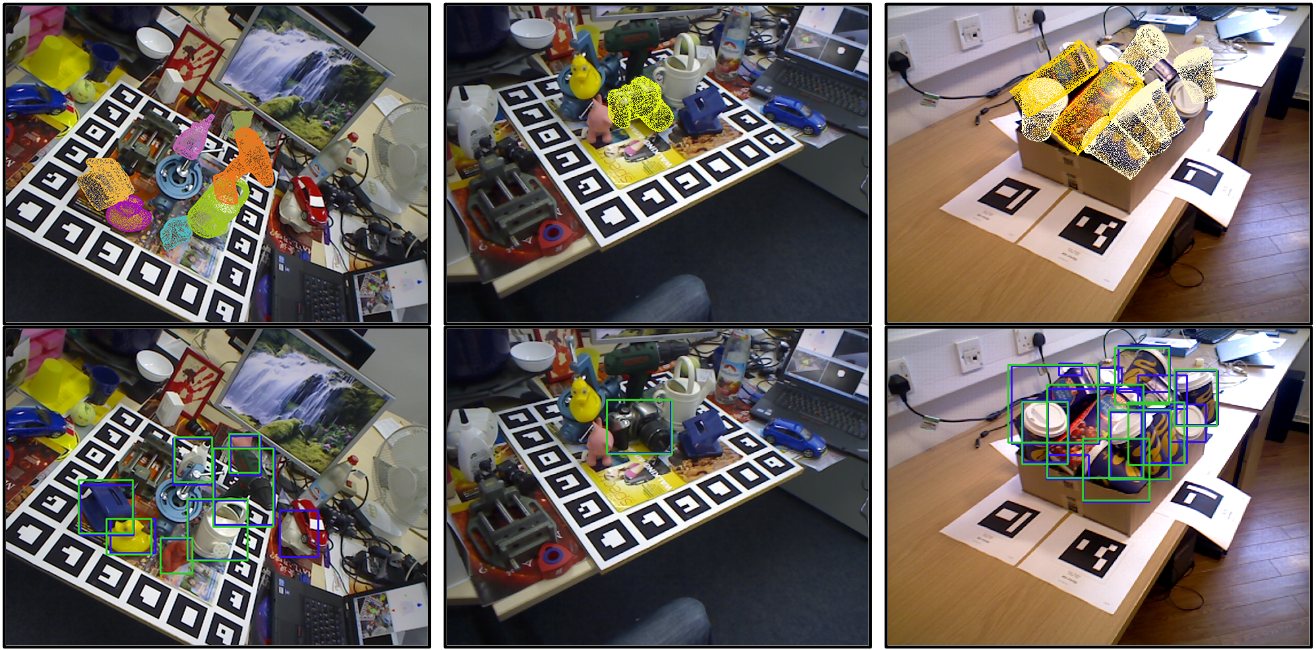}
   \caption{\textbf{Qualitative results on LM-O, LM and IC-BIN.} Top row shows reprojected object meshes based on the estimated poses, bottom row shows bounding box estimates. Blue boxes indicate the ground truth while green boxes indicate estimates.}
   \label{fig:qual}
   \vspace{-2ex}
\end{figure}

\section{Conclusion} \label{sec:conclusion}

This paper presented a framework for pose estimation that processes up to $70$ object instances in a single image with effectively constant runtime and up to $90$ in real-time ($>24$ fps).
Our end-to-end trainable, single-staged approach achieves up to $35$ times faster runtime than state-of-the-art approaches with similar formulation while also generating similar pose estimation accuracy.
Directly regressing the 6D pose from sparse intermediate geometric correspondences in the image space enables efficient network scaling with respect to the amount of object classes and instances. 
During test time, multiple instances are handled based on their 2D overlaps, which results in negligible runtime increase with respect to the number of instances. 
As such, we have developed an object pose estimator that is applicable and useful for a broad variety of real-time tasks.
In the future, we plan to improve our work by learning the intermediate representation in a self-supervised manner in order to overcome the necessity of defining geometric correspondences.


\section*{Acknowledgement}
We gratefully acknowledge the support of the EU-program EC Horizon 2020 for Research and Innovation under grant agreement No. 101017089, project TraceBot, the support by the Austrian Research Promotion Agency (FFG) under grant agreement No. 879878, project K4R, the Austrian Science Fund (FWF) under grant agreement No. I3969-N30, project InDex, and the NVIDIA Corporation for the donation of a GPU used for this research.

{\small
\bibliographystyle{ieee_fullname}
\bibliography{egbib}
}


\section*{Appendix}
\addcontentsline{toc}{section}{Appendices}
\renewcommand{\thesubsection}{\Alph{subsection}}

This supplementary manuscript provides the reader with an overview of the network layout in Section~\ref{sec:net}, additional training details in Section~\ref{sec:train},
discussion of the performance in comparison to the state of the art in Section~\ref{sec:perf}, a highlight of the capabilities of COPE in Section~\ref{sec:high} and concludes with a presentation of error cases in Section~\ref{sec:errors}.


\subsection{Network Design}\label{sec:net}

ResNet101~\cite{he2016deep} in conjunction with PFPN~\cite{thalhammer_pyrapose} is used for multi-scale feature extraction.
This is followed by three modules, one for location classification, bounding box estimation and keypoint estimation that are shared over feature pyramid levels.
Each of the modules consist of four convolution layers with Swish~\cite{mish} activation. The convolution layers of the classification module have 256 feature channels and those of the bounding box and the keypoint estimation module have 512 feature channels. 
Linearly activated output layers convolve the feature maps to a one-hot encoding for the object class, bounding box corners and the number of keypoints.
Feature maps are not spatially reduced when passing through these modules.
The estimated keypoints are fed to a module that directly learns 6D pose estimation, consisting of two convolution layers with 512 and 256 feature channels with Mish activation as well as a linearly activated output convolution with $a\cdot9$ output parameters, where $a$ is the number of classes.
Thus, the outputs are $3$ for translation and $6$ for rotation for each dataset object separately.
The parameter $d$ in Equation 2 of the submitted manuscript is set to $3$ in all experiments.

\subsection{Training Details}\label{sec:train}

\subsubsection*{Backbone}
The performance reduction occurring when estimating poses under domain shift is partially alleviated by setting low-level stages of the backbone to non-trainable~\cite{hodan2020epos,hinterstoisser_pretraining,zakharov2019dpod}.
EPOS~\cite{hodan2020epos} freezes the majority of the backbone, i.e. early and middle flow of Xception-65~\cite{chollet2017xception} when training exclusively on synthetic data.
We experienced this strategy to be infeasible for feature pyramid-based approaches since-prediction are made from different feature map resolutions taken also from early and intermediate feature maps of the backbone. 
Thus, freezing the weights of all layers up to the output layers reduces the pose estimation performance since feature learning for the present task is limited.
Hence, further investigating the adaption of the backbone to have deeper early stages might lead to improved domain transfer.

\subsubsection*{Hyperparameters}
All results are presented with the same set of hyperparameters, the only exception is Table 2 in the submitted manuscript. 
For the ablations, the parameter settings are mentioned in the table.

\begin{figure}[t!]
   \centering
   \includegraphics[width=\columnwidth]{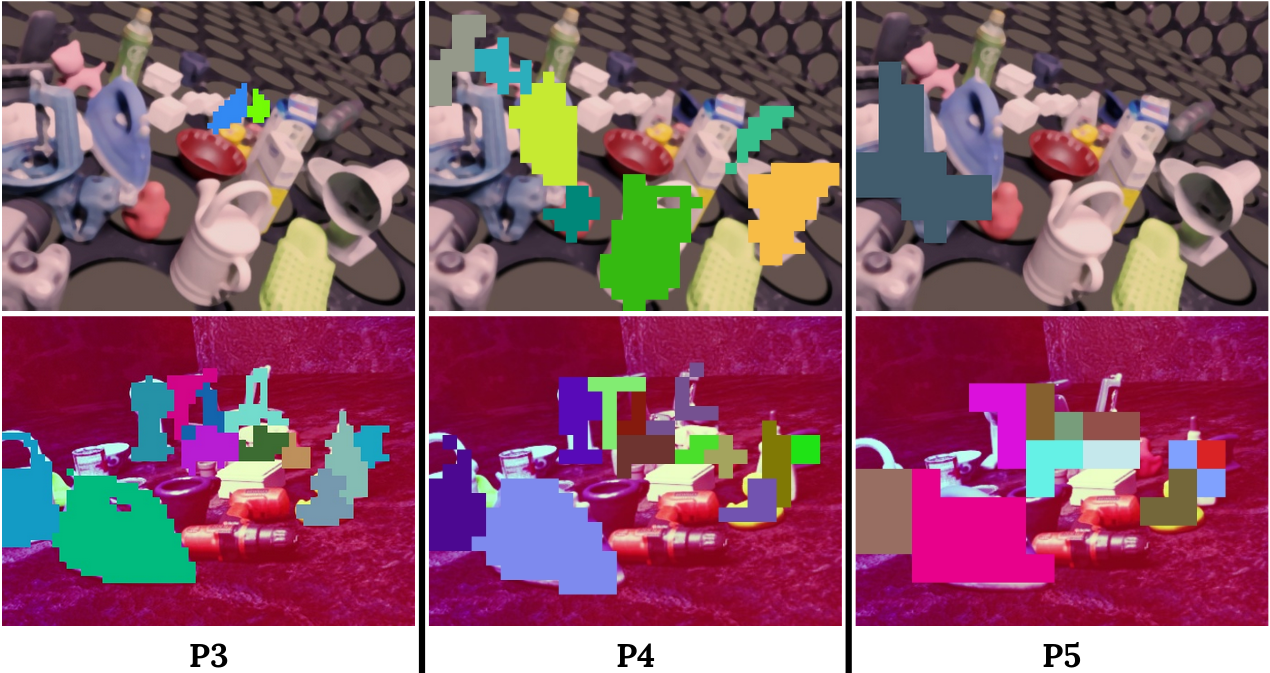}
   \caption{\textbf{Comparison of Training Location Sampling:} Utilizing the object dimensions for sampling the pyramid level to train on (\textbf{top}) leads to a similar amount of true locations per object and prevents ambiguous training locations.  Sampling small objects in coarse feature map resolutions results in aliasing effects that are detrimental to convergence (\textbf{bottom}). Images are cropped to improve visibility. Best viewed on a screen.}
   \label{fig:trainloc}
\end{figure}

\textbf{Training:} The relevant hyperparameters are, apart from the network layout and optimization itself, the base of the logarithm for location sampling, the visibility threshold for foreground samples and the color and image space augmentation hyperparameters.
A visual comparison for location sampling on physically-based rendering images of the LM~\cite{hinterstoisser2012model} objects is provided in Figure~\ref{fig:trainloc}.
The top row shows our physics-based scheme, the bottom shows using all pyramid levels for training.
Without providing quantitative analysis it is already visible that using all feature pyramid levels leads to aliasing effects in the lower resolved pyramid levels, while explicitly using object depth and dimension samples a similar amount of true locations per object.

\begin{table*}[ht!]
\caption{Color space augmentations applied during training.}
\label{tab:aug}
\begin{center}{
\begin{tabular}{c|c|c}
\hline
Augmentation & Chance (per channel) & Range\\
\hline
gaussian blur & 0.2 & $\sigma \sim \mathcal{U}(0.0, 2.0)$\\
average/median/motion blur & 0.2 & $\sigma \sim \mathcal{U}(3, 7)$ \\
bilateral blur & 0.2 & $\sigma \sim \mathcal{U}(1, 7)$ \\
hue/saturation & 0.5 & $\mathcal{U}(-15, 15)$  \\
grayscale & 0.5 & $\mathcal{U}(0.0, 0.2)$  \\
add & 0.5 (0.5) & $\mathcal{U}(-0.04, 0.04)$  \\
multiply & 0.5 (0.5) & $\mathcal{U}(0.75, 1.25)$  \\
gamma contrast & 0.5 (0.5) &  $\mathcal{U}(0.75, 1.25)$ \\
sigmoid contrast & 0.5 (0.5) &  $\mathcal{U}(0, 10)$  \\
logarithmic contrast & 0.5 (0.5) & $\mathcal{U}(0.75, 1.0)$  \\
linear contrast & 0.5 (0.5) & $\mathcal{U}(0.7, 1.3)$  \\
\hline
\end{tabular}
}
\end{center}
\end{table*}

Training on objects with too much occlusion is detrimental to pose estimation performance. 
To overcome this issue we set the threshold for foreground samples to $0.25$ of the object visibility when computing $L_{cls}$; all other losses are computed for objects with at least $0.5$ object visibility.
We apply affine color space transformations to improve the domain transfer, as also mentioned in the manuscript, parameters and ranges are provided in Table~\ref{tab:aug}.
Additionally we randomly scale training images by $5\%$ to improve translation equivariance of our trained models.

\textbf{Inference:} Parameters that require manual assignment are the detection threshold, the Intersection-over-Union (IoU) for clustering hypotheses, number of hypotheses to use for the pose computation per instance and the maximal number of instances per image.
All image locations with a detection threshold above $0.5$ are considered as foreground, thus as true locations containing objects of interest and are consequently used for hypotheses clustering and pruning.
An IoU of $0.5$ is used for clustering instance hypotheses. 
An ablation for the amount of hypotheses to derive the final pose is presented in Table 3 in the submitted manuscript.
The hyperparameter is set to $10$ for all experiments apart from those ablating its influence.
For the setting of the manuscript, the number of maximal instances to detect is set to $100$. 
Yet, despite there being little restrictions for that parameter, increasing this threshold contributes little to nothing since $100$ instances to detect per image is already a considerably large number.
After clustering and pruning hypotheses, those that are not supported by another hypothesis are discarded.

\subsubsection*{Evaluation Metrics}

Comparison to the state of the art is provided using the performance score of the BOP challenge~\cite{hodan2020bop}.
The deviation of the estimated pose $\hat{P}$ to the ground truth $P$ is projected to a scalar value using the average recall of three error metrics.
These are the Visual Surface Discrepancy, the Maximum Symmetry-Aware Surface Distance and the Maximum Symmetry-Aware Projection Distance:
\begin{multline}\label{eq:VSD}
    e_{VSD} = \underset{p\in \hat{V} \cup V}{avg} \begin{cases} 
    0 &\text{if $p \in \hat{V} \cap V \wedge | \hat{D}(p) - D (p)| < \tau$ }, \\
    1 &\text{otherwise}
    \end{cases} \\
    e_{MSSD} = \underset{s \in S_{i}}{min}\ \underset{m \in M_{i}}{max} ||\hat{P}m - Ps||_{2}, \\
    e_{MSPD} = \underset{s \in S_{i}}{min}\ \underset{m \in M_{i}}{max} ||proj_{3D\to2D}(\hat{P}m) \\ - proj_{3D\to2D}(Psm)||_{2},
\end{multline}
where $\hat{V}$ and $V$ are sets of image pixels; $\hat{D}$ and $D$ are distance maps and $\tau$ is a misalignment tolerance. Distance maps are rendered and compared to the distance map of the test image to derive $\hat{V}$ and $V$.
$S_{i}$ is a set of symmetry transformations that depend on the visual ambiguities of the object mesh. $M_{i}$ is a subset of the mesh vertices and $proj_{3D\to2D}(.)$ denotes the projection to the image space.
For each of these metrics the average recall ($AR$) is measured when comparing errors to multiple error thresholds (and $\tau$ in the case of $e_{VSD}$).
Results are then reported as the Average Recall: $AR = (AR_{VSD} + AR_{MSSD} + AR_{MSPD})/3$.

Ablations are evaluated using the ADD(-S) recall~\cite{hinterstoisser2012model}:
\begin{align}
    e_{ADD} &= \underset{m \in M_{i}}{avg} ||\hat{P}m - Pm||, \\
    e_{ADDS} &= \underset{m_{1} \in M_{i}}{avg}\ \underset{m_{2} \in M_{i}}{min} ||\hat{P}m_{1} - Pm_{2}||.
\end{align}
ADD measures the average deviation of models points using the corresponding point distance.
For objects exhibiting symmetric transformations, the ADD-S error, using the closest point distance, is calculated.
We report the fraction of poses below the commonly used error threshold of $10\%$ of the object diameter. 

Results for object detection are reported using the the mean Average Precision (mAP) of the Microsoft COCO object detection challenge~\cite{lin2014microsoft}.

\subsection{Comparing Detection Performance}\label{sec:perf}

The results reported in Table 2 in the submitted manuscript indicate that the detection performance of COPE is inferior to that of FCOS~\cite{tian_fcos}.
However, this conclusion has to be drawn with caution since FCOS only performs 2D Detection.
The network size of FCOS is $\sim50$ million parameters just for object detection while COPE additionally predict geometric correspondences and direct 6D poses with only $\sim17$ million parameters more. 
Additionally, FCOS uses an input image resolution with up to $1333$ pixels for the larger image side while COPE uses $640 \times 480$ input images.
Thus, COPE solves twice as many tasks with higher complexity from images with half of the input resolution.

\subsection{Highlights of COPE}\label{sec:high}

\begin{figure*}[ht]
   \centering
   \includegraphics[width=1.5\columnwidth]{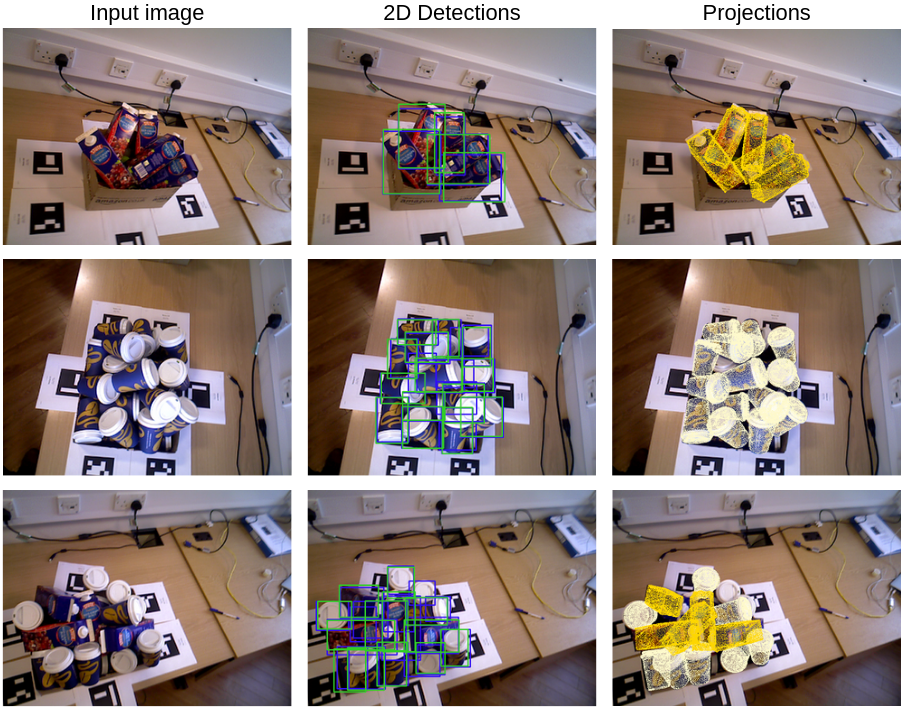}
   \caption{\textbf{Multiple Object Instances:} Columns are, from left to right, input image, 2D detections and reprojected object mehses based on the estimated poses. Each instance is indicated with a specific color. Green and blue bounding boxes correspond to estimates and ground truth, respective. Best viewed on screen.}
   \label{fig:highlightsICBIN}
\end{figure*}

\textbf{Handling Multiple Mutually Occluding Objects}
The top row of Figure~\ref{fig:highlightsICBIN} shows accurate bounding box and pose estimates on IC-BIN's~\cite{DBLP:journals/corr/DoumanoglouKMK15} \textit{Juice}. 
Due to the end-to-end trainability and the parallel learning of detection and pose estimation, COPE learns to effectively handle multiple mutually occluding instances of the same object.
Increased mutual occlusion of instances of \textit{Coffeecup} is displayed in the middle row, which again shows accurate bounding box and pose estimates for all the visible object instances.
The bottom row shows a scenario where both \textit{Juice} and \textit{Coffeecup} are present.
Ultimately, a false positive detection of \textit{Juice} occurs in the center of the bulk due to the heavy mutual occlusion of multiple instances.

\begin{figure*}[ht]
   \centering
   \includegraphics[width=1.5\columnwidth]{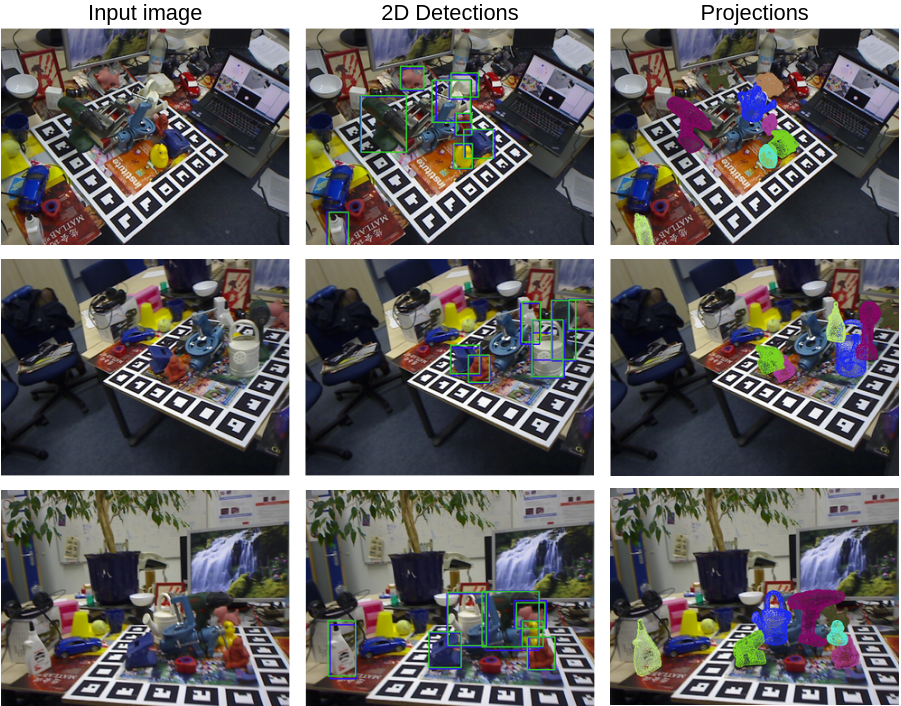}
   \caption{\textbf{Occlusion Handling on LM-O:} Columns are, from left to right, input image, 2D detections and reprojected object mehses based on the estimated poses. Each instance is indicated with a specific color. Green and blue bounding boxes correspond to estimates and ground truth, respective. Best viewed on screen.}
   \label{fig:highlightsLMO}
\end{figure*}

\textbf{Handling Occlusion in Clutter}
Figure~\ref{fig:highlightsLMO} displays accurate bounding box and pose estimates on occluded examples of LM-O's~\cite{brachmann2014learning} \textit{Ape}, \textit{Can} and \textit{Eggbox}.
The middle row shows similarly occluded examples of \textit{Drill}, \textit{Holepunch} and \textit{Glue} and the bottom row for \textit{Cat} and \textit{Duck}. 
Despite only training one model for all of LM's objects COPE is robustly handling each object, even under occlusion.

\subsection{Error Cases}\label{sec:errors}

\begin{figure*}[ht]
   \centering
   \includegraphics[width=1.5\columnwidth]{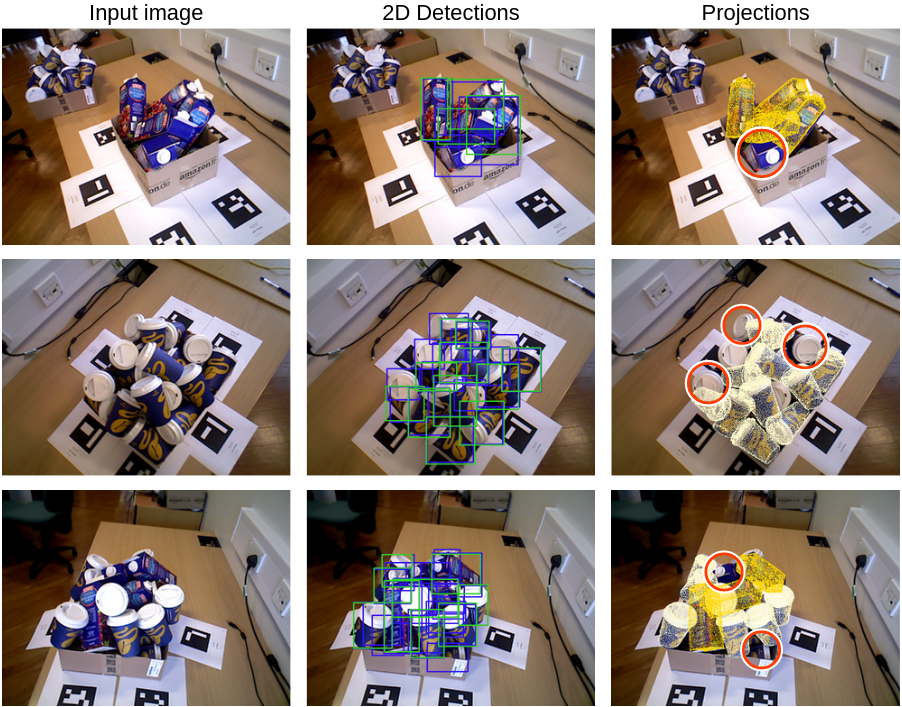}
   \caption{\textbf{Error Cases on IC-BIN:} Columns are, from left to right, input image, 2D detections and reprojected object mehses based on the estimated poses. Each instance is indicated with a specific color. Green and blue bounding boxes correspond to estimates and ground truth, respective. Errors are indicated with a red and white circle. Best viewed on screen.}
   \label{fig:errorICBIN}
\end{figure*}

Figure~\ref{fig:errorICBIN} presents recurring errors on IC-BIN.
The top row shows an instance of \textit{Juice} in top-view not being detected, indicated with a red and white circle.
Despite the high relative visibility of \textit{Juice} these reduced views are not often sampled during training data generation and are thus difficult to detect during runtime.
The middle row displays a similar error occuring for \textit{Coffeecup}, again indicated with red and white circles.
Multiple top-view orientated instances are not detected and thus result in false negative detections.
The bottom row shows one instance of each \textit{Coffeecup} and \textit{Juice} not being detected despite providing rich visual features.
Assigning true training locations on objects with too low visibility leads to reduced performance during inference.
As such we treat objects that are largely occluded, i.e. with less than $25\%$ relative object visibility, as background during training.
However, this leads to cases where discriminative object portions are visible in the image, yet are treated as background, as can be seen here.
Further investigation is required to consider the abundance of features during training target sampling to overcome such issues.

\begin{figure*}[ht]
   \centering
   \includegraphics[width=1.5\columnwidth]{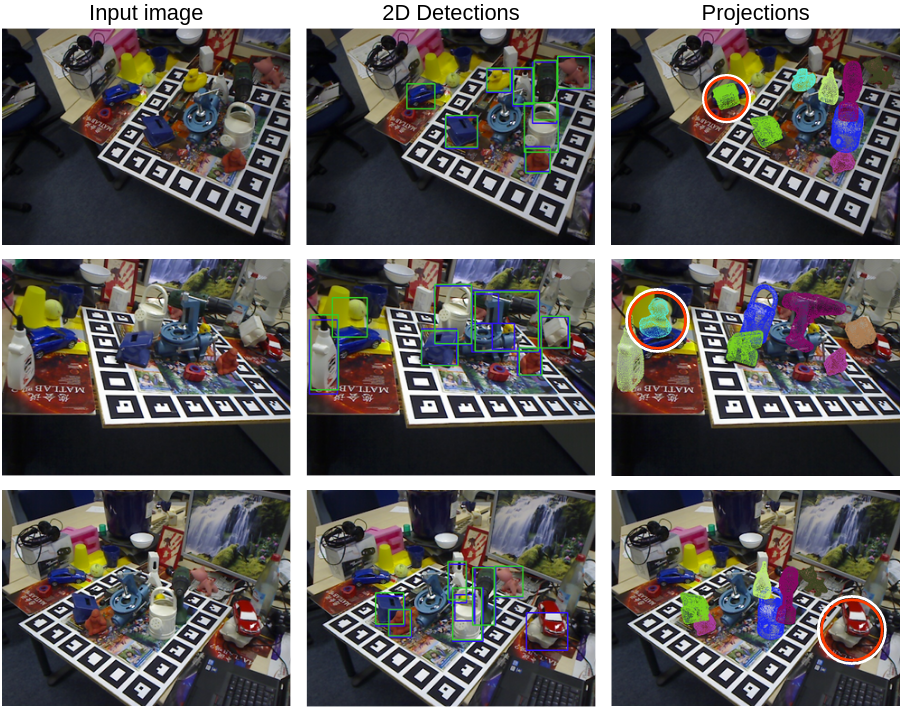}
   \caption{\textbf{Error Cases on LM-O:} Columns are, from left to right, input image, 2D detections and reprojected object mehses based on the estimated poses. Each instance is indicated with a specific color. Green and blue bounding boxes correspond to estimates and ground truth, respective. Errors are indicated with a red and white circle. Best viewed on screen.}
   \label{fig:errorLMO}
\end{figure*}

Figure~\ref{fig:errorLMO} presents common error cases on LM-O, again indicated with red and white circles.
The top row displays a false positive detection of the \textit{Holepunch} on a toy car with the same color and very similar material properties as the object of interest.
The middle row shows a similar false-positive detection of the object \textit{Duck}.
In the bottom row an example of \textit{Eggbox} with an occlusion pattern that is unlikely to be similarly sampled when randomizing object placements using physical modelling is displayed.
Since it is unlikely that objects of roughly the same size end up being stacked, these cases are not experienced during training~\cite{denninger}.

\end{document}